\documentclass[fleqn,twoside]{article}
\usepackage[headings]{espcrc2}
\usepackage{amsmath}

\readRCS
$Id: espcrc2.tex,v 1.2 2004/02/24 11:22:11 spepping Exp $
\usepackage{graphicx, balance}
\graphicspath{ {images/} }
\usepackage[figuresright]{rotating}
\usepackage{multirow}

\title{\textbf{Non Binary Local Gradient Contours for Face Recognition}}

\author{
Abdullah Gubbi\address[DCE]{Department of Electronics and Communication, P.A. College of Engnineering, Mangalore, Nadupadavu, Mangalore, India, Contact: abdullahgubbi@yahoo.com\\},
{Mohammad Fazle Azeem\address [DEE]{Senior  IEEE Member, Department of Electrical and Electronics Engineering, Aligarh Muslim University, India, Contact: mf.azeem@gmail.com\\}},
{M Sharmila Kumari\address[DCSE]{Department of Computer Science and Engineering, P A College of Engineering, Nadupadavu, Mangalore, India. Contact: sharmilabp@gmail.com \\}}
}

\runtitle{Non Binary Local Gradient Contours for Face Recognition}
\runauthor{Abdullah Gubbi, et al.,}

\begin{document}
\begin{abstract}
As the features from the traditional Local Binary patterns (LBP) and Local Directional Patterns (LDP) are found to be ineffective for face recognition, we have proposed a new approach derived on the basis of Information sets whereby the loss of information that occurs during the binarization is eliminated. The information sets expand the scope of fuzzy sets by connecting the attribute and the corresponding membership function value as a product. Since face is having smooth texture in a limited area, the extracted features must be highly discernible. To limit the number of features, we consider only the non overlapping windows. By the application of the information set theory we can reduce the number of feature of an image. The derived features are shown to work fairly well over eigenface, fisherface and LBP methods.  \\\\

{\bf Keywords:} Local Binary Pattern, Local Directional Pattern, Information Sets, Gradient Contour, Support Vector Machine, KNN, Face Recognition.
\end{abstract}

\maketitle
\section{INTRODUCTION}
In face recognition, the major issue to be addressed is the extraction of features which are discriminating in nature \cite{1}, \cite{2}. The accuracy of classification depends upon which texture feature of the face are extracted e.g., geometrical, statistical, local or global features in addition to representation of these features and the design of corresponding classifier. Normally, the feature extraction algorithm should produce little variance of features within the class and large variance between the classes. There are typically two common approaches to extract facial features: geometric-feature-based and appearance-based methods. The geometric-feature-based [\cite{3}, \cite{4}] method encodes the shape and locations of different facial components, which are combined into a feature vector that represents the face. An illustration of this method is the graph-based method \cite{5}, that uses several facial components to create a representation of the face and process it. The Local-Global Graph algorithm \cite{5} approach makes use Voronoi tessellation and Delaunay graphs to segment local features and builds a graph. These features are combined  into a local graph, and then the skeleton (global graph) is created by interrelating the local graphs to represent the topology of the face. The major requirements of geometric-feature-based methods is accurate and reliable facial feature detection and tracking, which is difficult to accommodate in many situations. In the case of appearance based methods, there are many methods for the holistic classes such as, Eigenfaces \cite{6} and Fisherfaces \cite{7}, which are built on Principal Component Analysis (PCA) \cite{6}, to the more recent 2D-PCA \cite{8}, and Linear Discriminant Analysis \cite{9} are also examples of holistic methods.  The \cite{10} and \cite{11} makes use of image filters, either on the whole face to create holistic features, or some specific face-region to create local features, to extract the appearance changes in the face image. The recognition rate of the appearance-based methods is good in constrained environment but their recognition rate deteriorates in environmental variation \cite{12}. Rui Huang et. al., have proposed a hybrid face recognition method that combines appearance and local feature analysis-based scheme  using a Markov Random Field (MRF). The face images are divided into small patches, and the MRF model is used to represent the relationship between the image patches and the patch ID's. The MRF model is first learned from the training image patches, given a test image. The most probable patch ID's are then inferred using the belief propagation (BP) hybrid model. Finally, the ID of the image is determined by a voting scheme from the estimated patch ID's. \\
Although these methods have been studied widely, local descriptors have gained attention because of their robustness to illumination and pose variations. Lu et.al., \cite{13} have presented a method for face recognition based on parallel neural networks and fuzzy clustering. The face patterns are divided into several small-scale neural networks based on fuzzy clustering and they are combined to obtain the recognition results. In \cite{14}, the authors have presented a method for face recognition combining modular neural networks and two interval type-2 fuzzy inference systems (FIS-2) for face recognition. The first (FIS-2) is used for edge detection in the training data, and the second one to find the ideal parameters for the Sugeno integral as a decision operator.\\
The theory of Information sets and Information processing developed to expand the scope of fuzzy sets and fuzzy modeling \cite{15} in bio-metrics. The two apparent drawbacks of fuzzy sets such as treating the information source (attributes) and their membership values separately in all the problems dealing with the fuzzy logic theory; and not accounting for probability (the occurrence) of information sources are the motivation for the present work. The membership function that gives the degree of association of a particular information source value in fuzzy sets is always considered ignoring the information source value. In reality, the information is a combination of information source and an agent’s opinion about it (membership function).  An attempt is made to overcome these drawbacks in the present work  by introducing the concept of information set that represents the amount of information or uncertainty in an information source.\\ 
Motivation: The problem with many of the pattern recognition  system is the processing the binarized images that results in loss of information source. Especially, the LBP \cite{16} has a problem with process of encoding. The LBP encodes the local neighborhood intensity by using the centre pixel as a threshold for a sparse sample of the neighboring pixels. The few number of pixels used in this method introduce several problems: Firstly, it limits the accuracy of the method. Secondly, the method discards most of the information in the neighborhood. Moreover, most of the methods we saw in the literature were partition the  image into small segments and restricting the bins for histogram construction followed by  histogram concatenation which again leads to loss of information as the neighboring feature values are assigned to same bins. \\
The remaining part of the paper is organized as follows. In section 2, we present some basics of information set and information processing. The section 3 is devoted to describe the family of texture descriptors based on information set. In section 4, we present the proposed methodology of feature extraction based on the theoretical aspects presented in section 3. The classification based on SVM and KNN classifier is presented in section 5. Experimental results are shown in section 6 followed by comparative analysis in section 7 and section 8 summarizes the conclusions that are drawn from our work.
\section{INFORMATION SETS AND INFORMATION PROCESSING}
Consider a fuzzy set constituted by a number of gray levels I=$\left\{I_{ij}\right\}$ in a window. When these are fitted with a membership function, their membership function values denoted by $\left\{\mu_{ij}\right\}$ represent the degree of association of gray levels to the set. Thus, the pair formed by the gray level and the membership function is considered as part of a fuzzy set. However, when the fuzzy set is considered in the context of the information theoretic entropy called Hanman-Anirban entropy  \cite{17}, the pair becomes a product leading to the concept of the Information set. The proposed information theoretic methods originate from Hanman-Anirban entropy function when the probabilities are replaced by the information sources and the assumption that the sum of information sources is equal to unity is relaxed. With this, the resolution of the function improves leading to increased distinguish ability such as when the details of an image become more discernible as its size increases. \\
The multimedia components (i.e. an image, speech, text or video) after granualization are considered as the information sources. The granularization amounts to partitioning in the case of an image. The property values, attributes or cues comprising the information sources contained in windows or frames form the fuzzy sets. The distribution of these information sources in the fuzzy sets require an appropriate membership function. Let us consider the commonly used membership functions such as the exponential and Gaussian type functions. \\
The Exponential and Gaussian type membership functions are given by:
\begin{equation}
\mu_{ij}^{e}=e^{-\left \{ \frac{\left | I_{ij}-I(ref) \right \vert}{f_{h}^{2}} \right \}} 
\end{equation}
\begin{equation}
\mu_{ij}^{g}=e^{{-\left \lbrack \frac{ I_{ij}-I(ref)}{\sqrt{2}f_{h}} \right \rbrack}^2}
\end{equation}

The fuzzifier $\left\{f_{h}^{2}\right\}$ in (1) is devised by Hanmandlu et al. in \cite{17} and it gives the spread of attribute values with respect to the chosen reference (symbolized as \textit{ref}). It is defined as

\begin{equation}
f_{h}^{2}= \frac{\sum_{i=1}^{w} \sum_{j=1}^{w} (I(ref)-I_{ij})^{4}}{\sum_{i=1}^{w} \sum_{j=1}^{w} (I(ref)-I_{ij})^{2}}
\end{equation}

One can take $I(ref)$ as $I_{avg}$ or $I_{max}$ or $I_{min}$ from the values in a window, where $I_{avg}$,$I_{max}$ and $I_{min}$ are the average, maximum and minimum  of  the window respectively. It may be noted that the above fuzzifier gives more spread than is possible with variance as used in the Gaussian function.

\section{NON BINARY GRADIENT CONTOURS}
Here, we define the non binary gradient contour of a 3x3 gray scale image patch.

\begin{figure}
\centering
\includegraphics[width=60mm,scale=0.5]{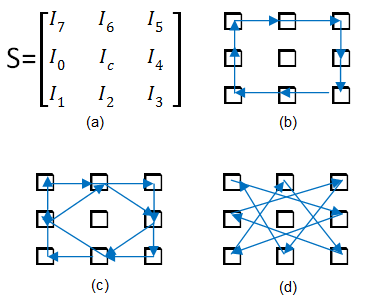}
\caption{(a) Spatial arrangement of a 3x3 gray scale pattern and schematic representation of the texture models considered in this paper,(b)single-loop, (c) double-loop, and(d)triple-loop versions of the non binary gradient contour concept.}
\label{Fig_1}
\end{figure}

Procedure: (1) The gradient between pairs of pixels is computed along a closed path around the central pixel S and 
(2) The gradients are not binarized because we lose the information during process of binarization. We propose to define the closed path in three different ways, namely: single-loop, double-loop and triple-loop, as shown in Fig. ~\ref{Fig_1}. Let S be a matrix representing the pixel intensities of a generic square neighborhood with a support of 3x3 region. Let $I_{c}$ be the gray-level of the central pixel and $I_{j}$ the gray-levels of the peripheral pixels (j $\in [1, 2, 3, \cdots ,7])$ which are arranged as follows.\\

The corresponding versions of the non binary gradient contours can accordingly be expressed as
\begin{eqnarray}
G_1 = [\vert \xi \left( I_{7}-I_{0}\right) \vert + \vert \xi\left(I_{6}-I_{7}\right) \vert + \nonumber  \\ 
\vert \xi \left(I_{5}-I_{6} \right) \vert + \vert \xi \left(I_{4}-I_{5}\right) \vert + \nonumber \\
 \vert \xi \left( I_{3}-I_{4}\right) \vert +  \vert \xi \left( I_{2}-I_{3}\right) +  \nonumber  \\
 \vert \xi \left( I_{1}-I_{2}\right) \vert + \vert \xi \left( I_{0}-I_{1}\right)]\\ \nonumber 
\end{eqnarray}

\begin{eqnarray}
G_{20} &= [\vert \xi \left( I_{6}-I_{0}\right) + \vert  \xi \left( I_{4}-I_{6}\right) \vert + \nonumber \\ 
&\qquad \vert \xi \left( I_{2}-I_{4}\right) \vert+ \vert \xi \left( I_{0}-I_{2}\right)\vert]
\end{eqnarray}


\begin{eqnarray}
G_{21} &= [\vert \xi \left( I_{7}-I_{1}\right) \vert + \vert \xi \left( ... I_{5}-I_{7}\right) \vert + \nonumber \\ &\qquad \vert \xi \left( I_{3} ...
-I_{5}\right) \vert+ \vert \xi \left( I_{1}-I_{3}\right) \vert ] ... 
\end{eqnarray}


\begin{equation}
G_2= \left[ G_{20}+G_{21} \right] 
\end{equation}


\begin{align}
G_3 &= [ \vert \xi \left( I_{5}-I_{0}\right) \vert + \vert \xi \left( I_{2}-I_{5}\right) \vert + \nonumber \\ &\qquad \vert \xi \left( I_{7}-I_{2}\right) \vert+ \vert \xi \left( I_{4}-I_{7}\right) \vert + \nonumber \\ &\qquad \vert \xi \left( I_{1}-I_{4}\right) \vert + \vert \xi \left( I_{6}-I_{1}\right) \vert + \nonumber \\ &\qquad \vert \xi \left( I_{3}-I_{6}\right) \vert + \vert \xi \left( I_{0}-I_{3}\right) \vert ]
\end{align}

It is useful to notice that the 8-tuples and 4-tuples above are actually functions of S, but the explicit dependency on S has been dropped from Eqs. $(4)-(8)$ to alleviate notation.                                           

To enhance the image classification information and improve the recognition accuracy, here we propose to use entropy information based on the Shannon entropy \cite{18}. The entropy is usually used to describe the information contained in the system. Shannon defined the concept of information and described the essential components of a communication system includes a sender, a receiver, a communication channel, and an encoding of the information set \cite{18}. The Shannon entropy can also be computed for an image, where the probabilities of the gray level distributions are considered in the Shannon Entropy formula. An image consisting of a single intensity will have a low entropy value; it contains very little information.\\
A high entropy value will be yielded by an image, which has much different intensity. In this manner, the Shannon entropy is also a measure of dispersion of a probability distribution. A distribution with a single sharp peak corresponds to a low entropy value, whereas a dispersed distribution yields a high entropy value. \\
The set of features used to describe face image can be regarded as alphabets of symbols, in analogy with a discrete noiseless channel \cite{19}. It has been observed that the dimension of the feature space represents a theoretical limit to the amount of information that can be conveyed through a face model. It is well known that the highest efficiency attainable by an alphabet occurs when its symbols are equally likely.\\
The proposed feature extraction algorithm is based on extracting features from the windows which are non overlapping and membership value is computed based on central pixel and Fuzzifier precisely from high informative local zones of the face image instead of utilizing the entire image in the form of window. The information content of different regions of a face image varies widely. It can be shown that, if an image of a face were divided into windows, all the windows may contain the same amount of information. It is expected that a close neighborhood of eyes, nose and lips contains more information than that possessed by the other regions of a  face image.\\

\section{PROPOSED METHOD}
\begin{itemize}
  \item Normalization of an image: We propose to use unit normalization to eliminate the problem of illumination and intensity variations. 
  \item Image Partitioning: We propose to partition the image into a non-overlapping blocks of size 3x3. The image is resized to obtain desired number of blocks which is optional.
  \item Membership computation: We compute the fuzzy membership of the central pixel (pixel under processing) and encodes the information within a window. In fact, we tend to lose the information about central pixel as in the case of LBP, LDP \cite{26} or any gradient pattern based approaches.
  \item Compute $f_{h}$ using equation 3.
  \item Membership value is given by 
  \begin{equation}
  \mu _{w}=\frac{i_{c}}{f_{h}}
  \end{equation}
  \item We compute $G_1$ which is in the non-binary form. We take absolute value of difference and add them up. As shown in equation 4, the non absolute values are most of the time are zero or nearly zero. This is due to that fact that the face has got smooth texture and values are very close to each other (the variance is very small)
  \item Computation of Shannon entropy: We propose to compute the Shannon entropy of each window as a feature value of a block  i.e.,  \begin{equation}
  F_{w}=-\mu _{w}G_{1}logG_{1}
  \end{equation}
  \item The features for each block of an image are concatenated to construct a feature vector.  Compute $F_{w}$ for complete image say if we have image of size 63x63, we get 441 features.
\end{itemize}
The similar procedure is adopted for $G_2$ and $G_3$.\\
The above process is repeated for all the images considered for training to create knowledge base. 

\section{Classification}

We propose here two ways of classification, one by having SVM \cite{20} as a trainer and other by using KNN classifier. In addition, we propose to use a new similarity measure instead of conventional Euclidean distance measure in the case of KNN classifier. The details are presented below.
\subsection{SVM based classification:} We divide the entire database into training set and testing set and apply Support Vector Machine. The SVM is a well-established machine learning approach which has been successfully adopted in various data classification problems. The concept of SVM is based on the modern statistical learning theory. For data classification, SVM firstly implicitly maps the data into a higher dimensional feature space and then constructs a hyper plane in such a way that the separating margin between the samples of different classes is optimal. This separating hyper plane then function as the decision surface. 
\subsection{KNN based classification:} The k-Nearest Neighbour classifier is amongst the simplest of all the machine learning methods. The principle of nearest neighbour classification was first proposed by Skellam \cite{21}. It is a non-parametric method for classifying objects. The classification is done based on how much close the test feature vector is to the training feature vectors in the feature space. An object is classified based on the majority votes of its neighbors. If k = 1, then the object is simply assigned to the class of its nearest neighbor We have not considered Euclidean distance due to the following reasons.\\
\begin{itemize}
  \item The Euclidean distance measure suffers whenever there is a high noise-to-signal ratio and negative spikes values and hence any correlation is difficult to establish. 
  \item The Euclidean distance measures the correlation between quantitative, continuous variables. It is not suitable for ordinal data, where preferences are listed according to rank instead of according to actual values.
\end{itemize}
Hence, We propose the new distance measure
\begin{equation}
d_{m}=\sum_{i=1}^{k}ln(1+\left | tr{_{i}} - te{_{i}} \right |)
\end{equation}
where $d{_{m}}$ is the distance measure, k is the dimension of the feature vector, $tr{_{i}}$  represents the $i{^{th}}$ element of training pattern. Similarly $te{_{i}}$ represents the $i{^{th}}$ element of a test pattern. 

\section {EXPERIMENTAL RESULTS}
This section presents the experimental results on standard face data-sets such as ORL, Sheffield and YALE, which are used as benchmark data-sets to evaluate the suitability of proposed method for face recognition problems.
It shall be noticed that, in the real life scenario the received information is invariably pruned either by correcting or by normalizing. This is due to the fact that the information received by our senses are perceived differently as differing information values as the perception is different and also  depending on how much importance we attach to the source. Similar to fuzzy variables, the information values are also natural variables and hence normalization is done on each image for further processing.In all the three datasets, we normalize all of the gray (pixel) values by dividing them by the maximum gray value in that image. So the pixel values lies in the range 0 to 1 (also known as $I_{ij}$, the normalized information source values). But for simplicity and of imparting the discriminating power which is accrued by not scaling, we choose the basic information as H, which is the product of the information source values and its membership function values. This product mislead the readers that the information sets are no way different from the fuzzy sets.\\

\subsection{Experimentation on ORL dataset:}
\begin{figure}
\centering
\includegraphics[width=60mm,scale=0.5]{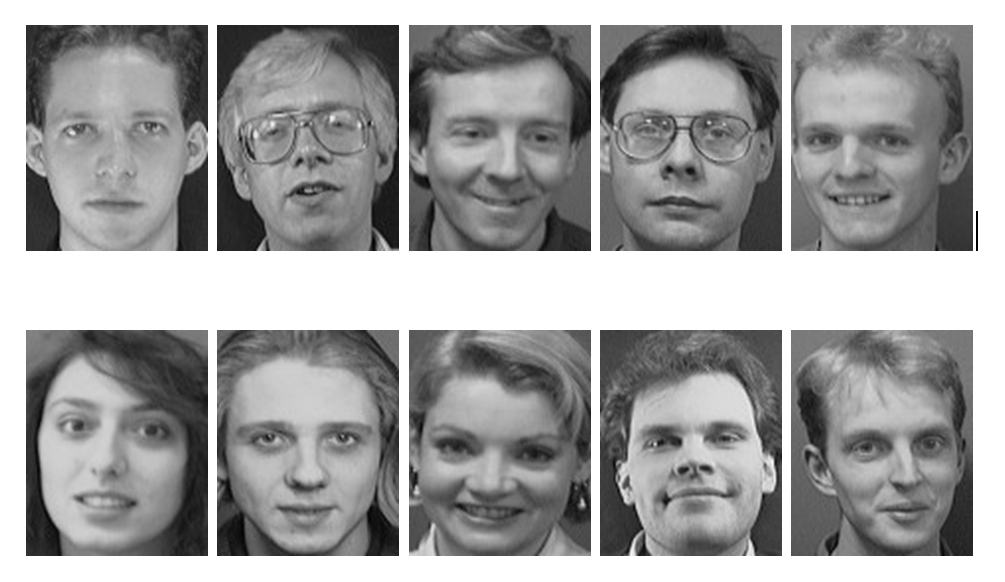}
\caption{Gray scale face images of some of the subject in ORL database.}
\label{Fig_2}
\end{figure}

The ORL dataset \cite{22} consist of images from 40 different individuals with 10 images of each person. All the face images were resized into images of 63x63 pixels to get 441 features. The datasets were partitioned into train and test set. The SVM classifier with a polynomial kernel of degree 1 and 2 is considered for training the classifier. The subset of some of the subjects of ORL dataset is shown in Fig. \ref{Fig_2}\\

The recognition accuracy of the proposed approach using SVM classifier and KNN classifier is reported in Tables \ref{Tab_1}, \ref{Tab_2}, and \ref{Tab_3} by respectively considering $G_1, G_2, G_3$.

\begin{table}[h]
\caption{Face Recognition accuracy on ORL dataset using NBLGC $G_2$.}
\label{Tab_2}
\begin{tabular}{|c|c|c|c|}
\hline
\multirow{2}{*}{\textbf{\begin{tabular}[c]{@{}c@{}}ORL  database\\ on NBLGC G2\end{tabular}}} & \multirow{2}{*}{\textbf{\begin{tabular}[c]{@{}c@{}}SVM\\ poly1\end{tabular}}} & \multirow{2}{*}{\textbf{\begin{tabular}[c]{@{}c@{}}SVM\\ poly2\end{tabular}}} & \multirow{2}{*}{\textbf{KNN}} \\
                            &                                                                                  &                                      &                               \\ \hline
\multirow{2}{*}{\begin{tabular}[c]{@{}c@{}}Training images7.\\ Testing images3.\end{tabular}}          & \multirow{2}{*}{\begin{tabular}[c]{@{}c@{}}95\%\end{tabular}}                                                            & \multirow{2}{*}{81.16\%}             & \multirow{2}{*}{\begin{tabular}[c]{@{}c@{}}93.33\%\end{tabular}}        \\
                            &                                                                                  &                                      &                               \\ \hline
\multirow{2}{*}{\begin{tabular}[c]{@{}c@{}}Training images6.\\ Testing images4.\end{tabular}}          & \multirow{2}{*}{\begin{tabular}[c]{@{}c@{}}91.25 \%\end{tabular}}                                                            & \multirow{2}{*}{\begin{tabular}[c]{@{}c@{}}88.12 \%\end{tabular}}             & \multirow{2}{*}{\begin{tabular}[c]{@{}c@{}}91.25 \%\end{tabular}}        \\
                            &                                                                                  &                                      &                               \\ \hline
\end{tabular}
\end{table}

\begin{table}[h]
\caption{Face Recognition accuracy on ORL dataset using NBLGC $G_2$.}
\label{Tab_2}
\begin{tabular}{|c|c|c|c|}
\hline
\multirow{2}{*}{\textbf{\begin{tabular}[c]{@{}c@{}}ORL  database\\ on NBLGC G2\end{tabular}}} & \multirow{2}{*}{\textbf{\begin{tabular}[c]{@{}c@{}}SVM\\ poly1\end{tabular}}} & \multirow{2}{*}{\textbf{\begin{tabular}[c]{@{}c@{}}SVM\\ poly2\end{tabular}}} & \multirow{2}{*}{\textbf{KNN}} \\
                            &                                                                                  &                                      &                               \\ \hline
\multirow{2}{*}{\begin{tabular}[c]{@{}c@{}}Training images7.\\ Testing images3.\end{tabular}}          & \multirow{2}{*}{\begin{tabular}[c]{@{}c@{}}93.33\%\end{tabular}}                                                            & \multirow{2}{*}{90\%}             & \multirow{2}{*}{\begin{tabular}[c]{@{}c@{}}93.33\%\end{tabular}}        \\
                            &                                                                                  &                                      &                               \\ \hline
\multirow{2}{*}{\begin{tabular}[c]{@{}c@{}}Training images6.\\ Testing images4.\end{tabular}}          & \multirow{2}{*}{\begin{tabular}[c]{@{}c@{}}92.5\%\end{tabular}}                                                            & \multirow{2}{*}{\begin{tabular}[c]{@{}c@{}}88.12\%\end{tabular}}             & \multirow{2}{*}{\begin{tabular}[c]{@{}c@{}}91.87\%\end{tabular}}        \\
                            &                                                                                  &                                      &                               \\ \hline
\end{tabular}
\end{table}

\begin{table}[h]
\caption{Face Recognition on ORL database using NBLGC $G_3$.}
\label{Tab_3}
\begin{tabular}{|c|c|c|c|}
\hline
\multirow{2}{*}{\textbf{\begin{tabular}[c]{@{}c@{}}ORL  database\\ on NBLGC G3\end{tabular}}} & \multirow{2}{*}{\textbf{\begin{tabular}[c]{@{}c@{}}SVM\\ poly1\end{tabular}}} & \multirow{2}{*}{\textbf{\begin{tabular}[c]{@{}c@{}}SVM\\ poly2\end{tabular}}} & \multirow{2}{*}{\textbf{KNN}} \\
                            &                                                                                  &                                      &                               \\ \hline
\multirow{2}{*}{\begin{tabular}[c]{@{}c@{}}Training images7.\\ Testing images3.\end{tabular}}          & \multirow{2}{*}{\begin{tabular}[c]{@{}c@{}}93.33\%\end{tabular}}                                                            & \multirow{2}{*}{\begin{tabular}[c]{@{}c@{}}88.33\%\end{tabular}}             & \multirow{2}{*}{\begin{tabular}[c]{@{}c@{}}93.33\%\end{tabular}}        \\
                            &                                                                                  &                                      &                               \\ \hline
\multirow{2}{*}{\begin{tabular}[c]{@{}c@{}}Training images6.\\ Testing images4.\end{tabular}}          & \multirow{2}{*}{\begin{tabular}[c]{@{}c@{}}91.25\%\end{tabular}}                                                            & \multirow{2}{*}{\begin{tabular}[c]{@{}c@{}}86.87\%\end{tabular}}             & \multirow{2}{*}{\begin{tabular}[c]{@{}c@{}}91.87\%\end{tabular}}        \\
                            &                                                                                  &                                      &                               \\ \hline
\end{tabular}
\end{table}

\begin{figure}
\centering
\includegraphics[width=73mm,scale=0.6]{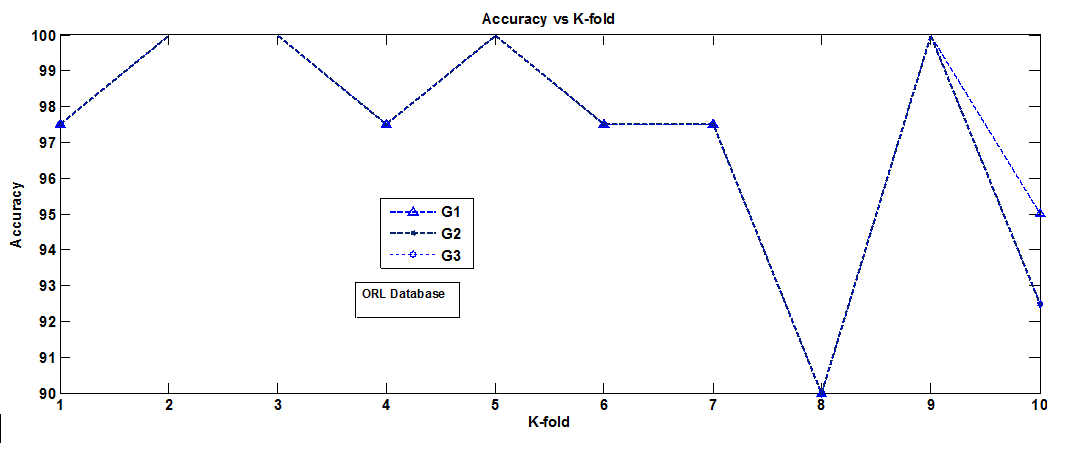}
\caption{K-fold accuracy on ORL dataset.}
\label{Fig_3}
\end{figure}

It is a well known fact that the k-fold cross validation is a common technique for estimating the performance of a classifier. The advantage of k-fold cross validation is that all the examples in the dataset are eventually used for both training and testing. Hence, we have conducted experimentation considering 10-fold cross validation process. The k-fold results considering SVM classifier varies from 90\% at lower end and 100\% at higher end for k varies from 1 to 10. The results are shown in the Fig. \ref{Fig_3}.

\begin{figure}
\centering
\includegraphics[width=73mm,scale=0.6]{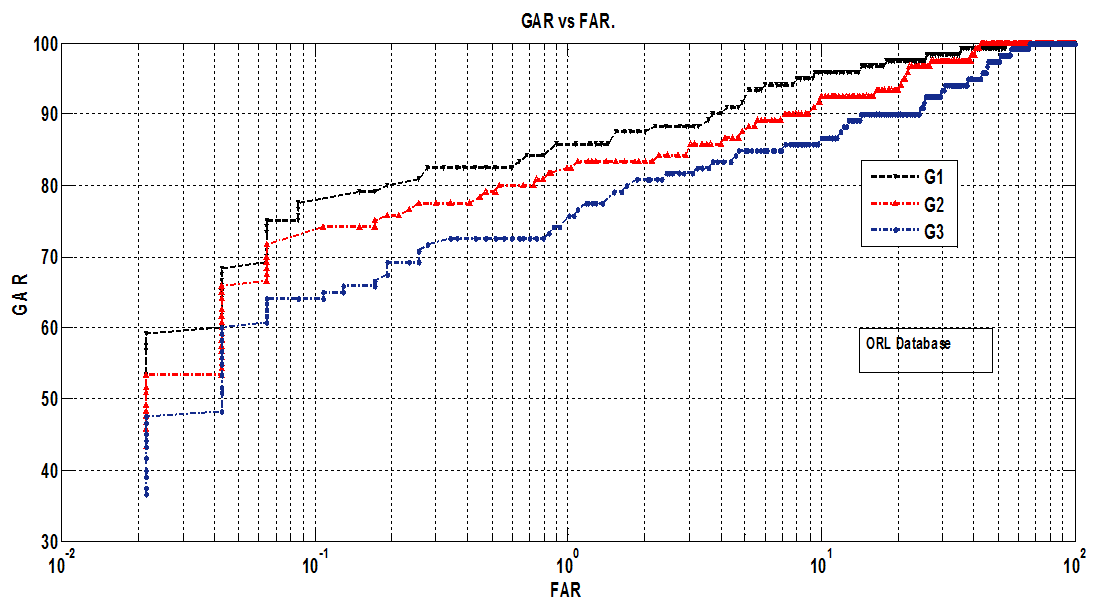}
\caption{The GAR Vs. FAR on ORL database with $G_1$, $G_2$ and $G_3$.}
\label{Fig_4}
\end{figure}

In addition, we have seen in the literature that most of the authors tested their approach and presented their results using ROC curves. We have presented the results of our approach on ORL dataset in Fig. \ref{Fig_4} considering the False Acceptance Rate (FAR) and Genuine Acceptance Rate (GAR). The FAR is the ratio of the number of successful fraud attempts to the total number of fraud attempts against a person and the GAR equals 100-FAR. Finally using these error rates, the receiver operating characteristics (ROC) which depicts the performance of an authentication system is plotted. The ROC plot is drawn FAR vs. GAR, with varying threshold values. From this plot, the threshold that yields the highest GAR corresponding to the lowest FAR need to be selected. It is evident from the graph that the $G_1$ outperform $G_2$ and $G_3$ for ORL databset.

\subsection{Experimentation on Sheffield dataset:}
\begin{figure}
\centering
\includegraphics[width=60mm,scale=0.5]{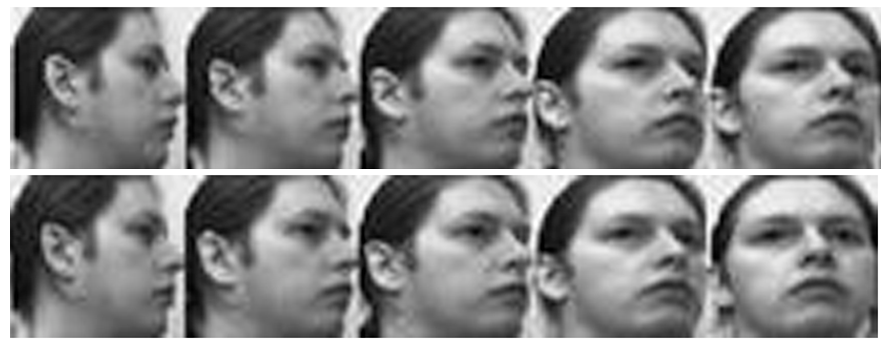}
\caption{Gray scale face images of single subject in Sheffield database.}
\label{Fig_5}
\end{figure}

The Sheffield (previously UMIST) Face Database consists of 564 images of 20 individuals (mixed race/gender/appearance). The Sheffield data base consist of 20 classes with minimum number of images per class as 23 and maximum of 58 images. Each individual is shown in a range of poses from profile to frontal views - each in a separate directory labeled $1_a,1_b,...1_t$ and images are numbered consecutively as they were taken. All the images  are in PGM format of approximately 220 x 220 pixels with 256-bit grey-scale. In this dataset, there is large orientation of face exist. The subset of some of the subjects of Sheffield dataset is shown in Fig. \ref{Fig_5}. Similar to previous experimentation, here too, the datasets were partitioned into train and test set. The SVM classifier with a polynomial kernel of degree 1 and 2 is considered for training the classifier. 

The recognition accuracy of the proposed approach on Sheffield database using SVM classifier and KNN classifier is reported in Tables \ref{Tab_4}, \ref{Tab_5} and \ref{Tab_6} by respectively considering $G_1, G_2, G_3$.

\begin{table}[h]
\caption{Face Recognition accuracy on Sheffield database using NBLGC $G_1$.}
\label{Tab_4}
\begin{tabular}{|c|c|c|c|}
\hline
\multirow{3}{*}{\textbf{\begin{tabular}[c]{@{}c@{}}Sheffield\\database\\
on NBLGC G1\end{tabular}}} & \multirow{3}{*}{\textbf{\begin{tabular}[c]{@{}c@{}}SVM\\ poly1\end{tabular}}} & \multirow{3}{*}{\textbf{\begin{tabular}[c]{@{}c@{}}SVM\\ poly2\end{tabular}}} & \multirow{3}{*}{\textbf{KNN}} \\
                            &                                                                                  &                                      &                               \\
                            &                                                                                  &                                      &                               \\ \hline
\multirow{2}{*}{\begin{tabular}[c]{@{}c@{}}Training images7.\\ Testing images21.\end{tabular}}          & \multirow{2}{*}{\begin{tabular}[c]{@{}c@{}}91.42\%\end{tabular}}     & \multirow{2}{*}{90\%}             & \multirow{2}{*}{90\%}        \\
                            &                                                                                  &                                      &                               \\ \hline
\multirow{2}{*}{\begin{tabular}[c]{@{}c@{}}Training images14.\\ Testing images14.\end{tabular}}          & \multirow{2}{*}{\begin{tabular}[c]{@{}c@{}}96.42\%\end{tabular}}     & \multirow{2}{*}{\begin{tabular}[c]{@{}c@{}}94.28\%\end{tabular}}             & \multirow{2}{*}{90\%}        \\
                            &                                                                                  &                                      &                               \\ \hline
\end{tabular}
\end{table}

\begin{table}[h]
\caption{Face Recognition accuracy on Sheffield database using NBLGC $G_2$.}
\label{Tab_5}
\begin{tabular}{|c|c|c|c|}
\hline
\multirow{3}{*}{\textbf{\begin{tabular}[c]{@{}c@{}}Sheffield\\database\\
on NBLGC G2\end{tabular}}} & \multirow{3}{*}{\textbf{\begin{tabular}[c]{@{}c@{}}SVM\\ poly1\end{tabular}}} & \multirow{3}{*}{\textbf{\begin{tabular}[c]{@{}c@{}}SVM\\ poly2\end{tabular}}} & \multirow{3}{*}{\textbf{KNN}} \\
                            &                                                                                  &                                      &                               \\
                            &                                                                                  &                                      &                               \\ \hline
\multirow{2}{*}{\begin{tabular}[c]{@{}c@{}}Training images7.\\ Testing images21.\end{tabular}}          & \multirow{2}{*}{\begin{tabular}[c]{@{}c@{}}90.95\%\end{tabular}}     & \multirow{2}{*}{\begin{tabular}[c]{@{}c@{}}87.61\%\end{tabular}}             & \multirow{2}{*}{\begin{tabular}[c]{@{}c@{}}90.47\%\end{tabular}}        \\
                            &                                                                                  &                                      &                               \\ \hline
\multirow{2}{*}{\begin{tabular}[c]{@{}c@{}}Training images14.\\ Testing images14.\end{tabular}}          & \multirow{2}{*}{\begin{tabular}[c]{@{}c@{}}97.14\%\end{tabular}}     & \multirow{2}{*}{\begin{tabular}[c]{@{}c@{}}94.28\%\end{tabular}}             & \multirow{2}{*}{\begin{tabular}[c]{@{}c@{}}90.71\%\end{tabular}}        \\
                            &                                                                                  &                                      &                               \\ \hline
\end{tabular}
\end{table}

\begin{table}[h]
\caption{Face Recognition accuracy on Sheffield database using NBLGC $G_3$.}
\label{Tab_6}
\begin{tabular}{|c|c|c|c|}
\hline
\multirow{3}{*}{\textbf{\begin{tabular}[c]{@{}c@{}}Sheffield\\database\\
on NBLGC G3\end{tabular}}} & \multirow{3}{*}{\textbf{\begin{tabular}[c]{@{}c@{}}SVM\\ poly1\end{tabular}}} & \multirow{3}{*}{\textbf{\begin{tabular}[c]{@{}c@{}}SVM\\ poly2\end{tabular}}} & \multirow{3}{*}{\textbf{KNN}} \\
                            &                                                                                  &                                      &                               \\
                            &                                                                                  &                                      &                               \\ \hline
\multirow{2}{*}{\begin{tabular}[c]{@{}c@{}}Training images7.\\ Testing images21.\end{tabular}}          & \multirow{2}{*}{\begin{tabular}[c]{@{}c@{}}90.47\%\end{tabular}}     & \multirow{2}{*}{\begin{tabular}[c]{@{}c@{}}86.66\%\end{tabular}}             & \multirow{2}{*}{\begin{tabular}[c]{@{}c@{}}89.52\%\end{tabular}}        \\
                            &                                                                                  &                                      &                               \\ \hline
\multirow{2}{*}{\begin{tabular}[c]{@{}c@{}}Training images14.\\ Testing images14.\end{tabular}}          & \multirow{2}{*}{\begin{tabular}[c]{@{}c@{}}96.42\%\end{tabular}}     & \multirow{2}{*}{\begin{tabular}[c]{@{}c@{}}94.28\%\end{tabular}}             & \multirow{2}{*}{\begin{tabular}[c]{@{}c@{}}91.42\%\end{tabular}}        \\
                            &                                                                                  &                                      &                               \\ \hline
\end{tabular}
\end{table}

\begin{figure}
\centering
\includegraphics[width=73mm,scale=0.6]{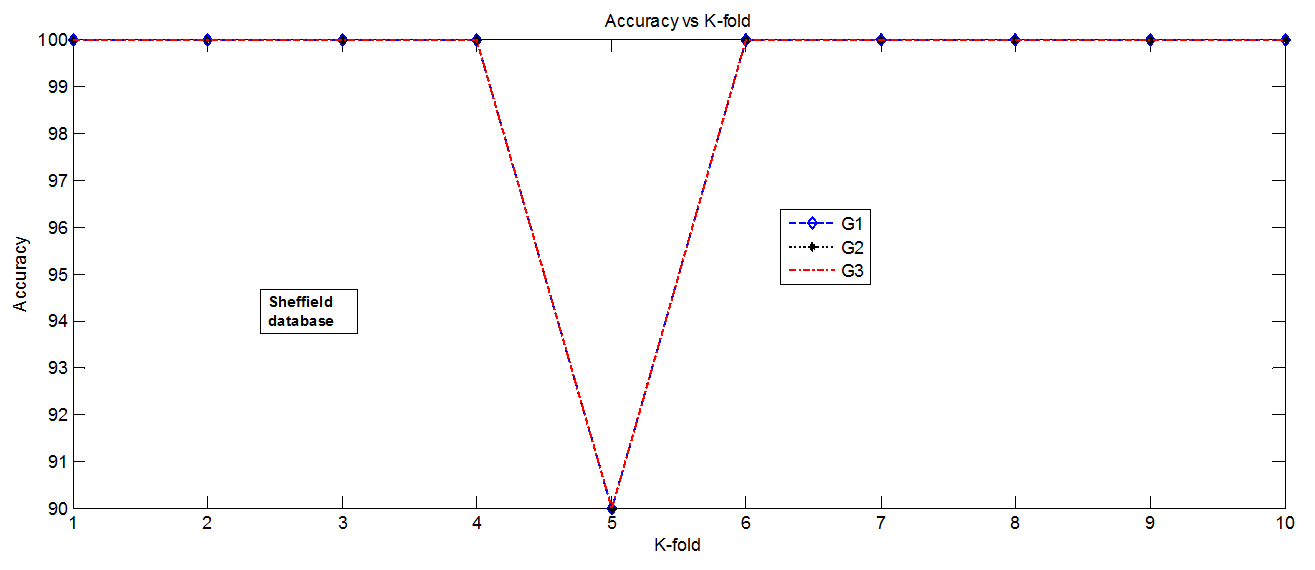}
\caption{K-fold accuracy on Sheffield database.}
\label{Fig_6}
\end{figure}

The experiments conducted on Sheffield database considering 10-fold cross validation process is given here. The k-fold results, considering SVM classifier, varies from 98\% at lower end and 100\% at higher end for k varies from 1 to 10. The results are shown in the Fig. \ref{Fig_6}.

\begin{figure}
\centering
\includegraphics[width=73mm,scale=0.6]{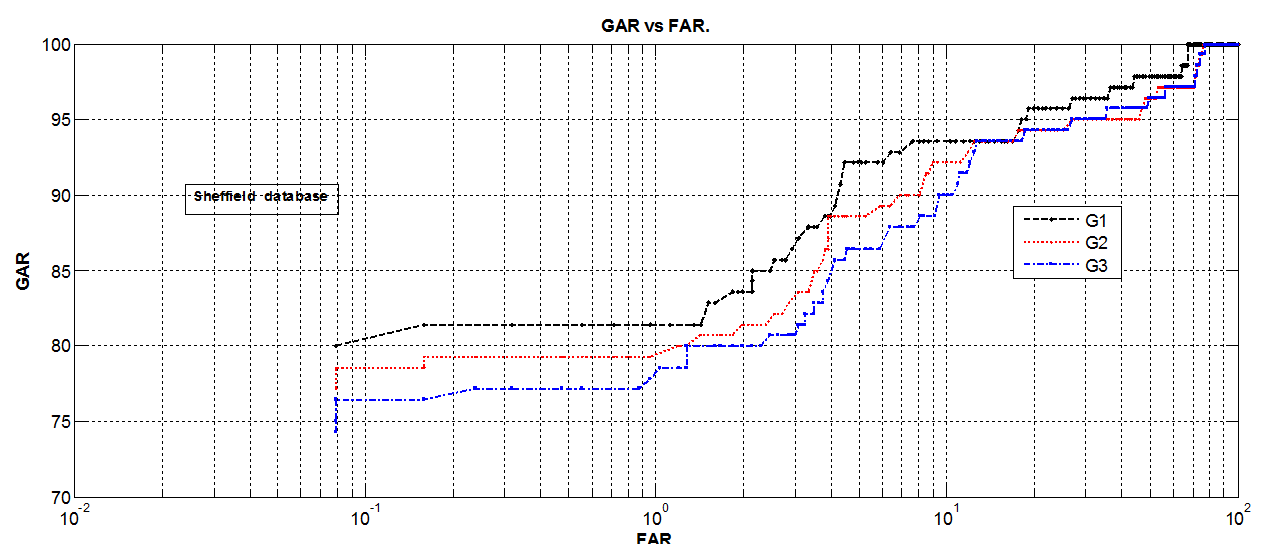}
\caption{The GAR Vs FAR on Sheffield database with $G_1$, $G_2$ and $G_3$.}
\label{Fig_7}
\end{figure}

The ROC plot is drawn FAR vs. GAR, with varying threshold values for Sheffield database also and shown in Fig. \ref{Fig_7}. From this plot, the threshold that yields the highest GAR corresponding to the lowest FAR need to be selected. It is evident from the graph that the $G_1$ outperform $G_2$ and $G_3$ for this databset.

\subsection{Experimentation on YALE dataset:}
The YALE dataset \cite{24} consists of 165 images of 15 individual. There are 11 images of each individual, one per different facial expression or configuration: center-light, with glasses, happy, left-light, without glasses, normal, right-light, sad, sleepy, surprised, and wink. The YALE database images have a fixed pose but different light illumination condition and expression. \\

\begin{figure}
\centering
\includegraphics[width=60mm,scale=0.5]{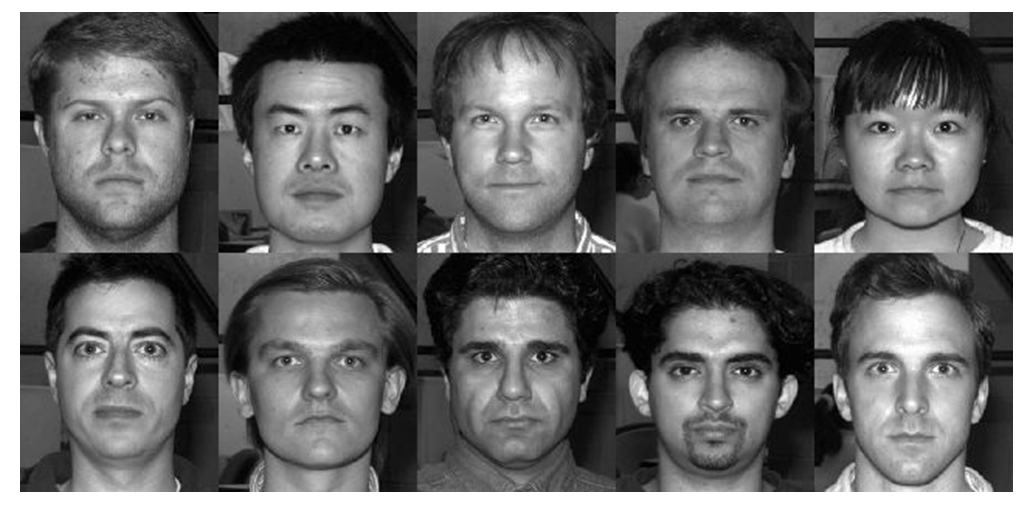}
\caption{Gray scale face images of some of the subject in YALE database}
\label{Fig_8}
\end{figure}

The subset of some of the subjects of YALE dataset is shown in Fig. \ref{Fig_8}. Similar to previous experimentation, here too, the datasets were partitioned into train and test set. The SVM classifier with a polynomial kernel of degree 1 and 2 is considered for training the classifier. 

The recognition accuracy of the proposed approach on YALE database using SVM classifier and KNN classifier is reported in Tables \ref{Tab_7}, \ref{Tab_8} and \ref{Tab_9} by respectively considering $G_1, G_2, G_3$.

\begin{table}[h]
\caption{Face Recognition accuracy on  YALE database using NBLGC $G_1$.}
\label{Tab_7}
\begin{tabular}{|c|c|c|c|}
\hline
\multirow{2}{*}{\textbf{\begin{tabular}[c]{@{}c@{}}YALE database \\ on NBLGC G1
\end{tabular}}} & \multirow{2}{*}{\textbf{\begin{tabular}[c]{@{}c@{}}SVM\\ poly1\end{tabular}}} & \multirow{2}{*}{\textbf{\begin{tabular}[c]{@{}c@{}}SVM\\ poly2\end{tabular}}} & \multirow{2}{*}{\textbf{KNN}} \\
                            &                                                                                  &                                      &                               \\ \hline
\multirow{2}{*}{\begin{tabular}[c]{@{}c@{}}Training images6.\\ Testing images5.\end{tabular}}          & \multirow{2}{*}{\begin{tabular}[c]{@{}c@{}}69.33\%\end{tabular}}                                                            & \multirow{2}{*}{68\%}             & \multirow{2}{*}{76\%}        \\
                            &                                                                                  &                                      &                               \\ \hline
\multirow{2}{*}{\begin{tabular}[c]{@{}c@{}}Training images5.\\ Testing images6.\end{tabular}}          & \multirow{2}{*}{\begin{tabular}[c]{@{}c@{}}66.66\%\end{tabular}}                                                            & \multirow{2}{*}{\begin{tabular}[c]{@{}c@{}}61.11\%\end{tabular}}             & \multirow{2}{*}{70\%}        \\
                            &                                                                                  &                                      &                               \\ \hline
\end{tabular}
\end{table}

\begin{table}[h]
\caption{Face Recognition accuracy on YALE database using NBLGC $G_2$.}
\label{Tab_8}
\begin{tabular}{|c|c|c|c|}
\hline
\multirow{2}{*}{\textbf{\begin{tabular}[c]{@{}c@{}}YALE database \\ on NBLGC G2
\end{tabular}}} & \multirow{2}{*}{\textbf{\begin{tabular}[c]{@{}c@{}}SVM\\ poly1\end{tabular}}} & \multirow{2}{*}{\textbf{\begin{tabular}[c]{@{}c@{}}SVM\\ poly2\end{tabular}}} & \multirow{2}{*}{\textbf{KNN}} \\
                            &                                                                                  &                                      &                               \\ \hline
\multirow{2}{*}{\begin{tabular}[c]{@{}c@{}}Training images6.\\ Testing images5.\end{tabular}}          & \multirow{2}{*}{\begin{tabular}[c]{@{}c@{}}74.66\%\end{tabular}}                                                            & \multirow{2}{*}{\begin{tabular}[c]{@{}c@{}}70.66\%\end{tabular}}             & \multirow{2}{*}{62.22\%}        \\
                            &                                                                                  &                                      &                               \\ \hline
\multirow{2}{*}{\begin{tabular}[c]{@{}c@{}}Training images5.\\ Testing images6.\end{tabular}}          & \multirow{2}{*}{\begin{tabular}[c]{@{}c@{}}63.33\%\end{tabular}}                                                            & \multirow{2}{*}{\begin{tabular}[c]{@{}c@{}}62.22\%\end{tabular}}             & \multirow{2}{*}{66.66\%}        \\
                            &                                                                                  &                                      &                               \\ \hline
\end{tabular}
\end{table}

\begin{table}[h]
\caption{Face Recognition accuracy on YALE database using NBLGC $G_3$.}
\label{Tab_9}
\begin{tabular}{|c|c|c|c|}
\hline
\multirow{2}{*}{\textbf{\begin{tabular}[c]{@{}c@{}}YALE database \\ on NBLGC G3
\end{tabular}}} & \multirow{2}{*}{\textbf{\begin{tabular}[c]{@{}c@{}}SVM\\ poly1\end{tabular}}} & \multirow{2}{*}{\textbf{\begin{tabular}[c]{@{}c@{}}SVM\\ poly2\end{tabular}}} & \multirow{2}{*}{\textbf{KNN}} \\
                            &                                                                                  &                                      &                               \\ \hline
\multirow{2}{*}{\begin{tabular}[c]{@{}c@{}}Training images6.\\ Testing images5.\end{tabular}}          & \multirow{2}{*}{72\%}                                                            & \multirow{2}{*}{72\%}             & \multirow{2}{*}{60\%}        \\
                            &                                                                                  &                                      &                               \\ \hline
\multirow{2}{*}{\begin{tabular}[c]{@{}c@{}}Training images5.\\ Testing images6.\end{tabular}}          & \multirow{2}{*}{64\%}                                                            & \multirow{2}{*}{61\%}             & \multirow{2}{*}{55.55\%}        \\
                            &                                                                                  &                                      &                               \\ \hline
\end{tabular}
\end{table}

\begin{figure}
\centering
\includegraphics[width=73mm,scale=0.6]{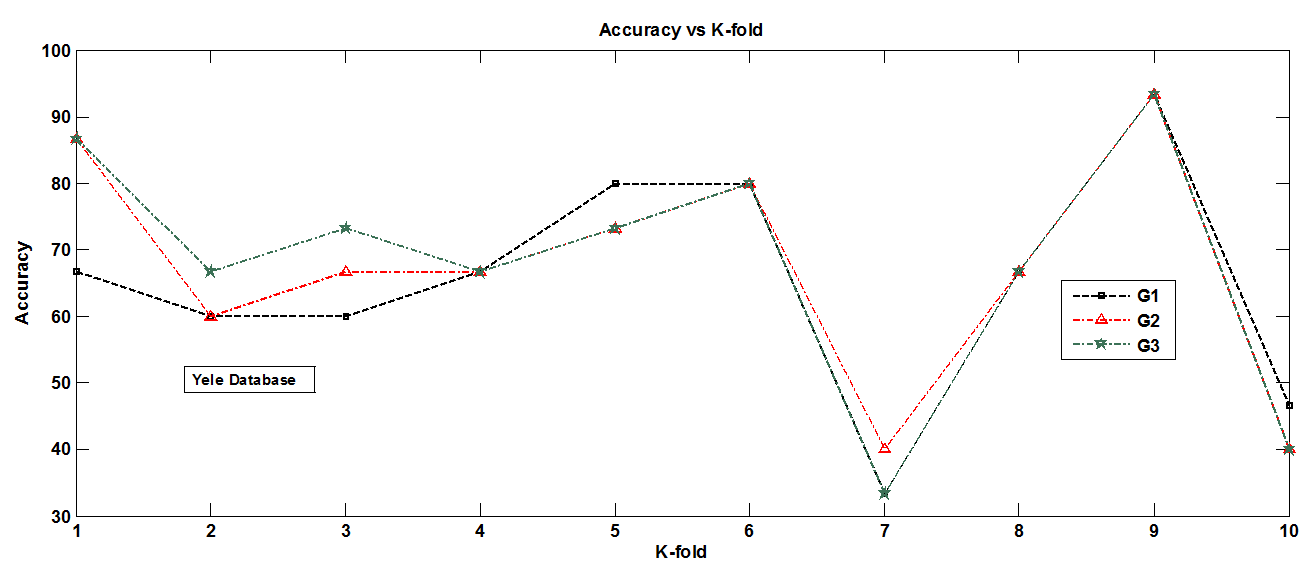}
\caption{K-fold accuracy on YALE database.}
\label{Fig_9}
\end{figure}

The experimentation conducted on YALE database considering 10-fold cross validation process is given here. The k-fold results considering SVM classifier varies from 32\% at lower end and 95\% at higher end for k varies from 1 to 10. The results are shown in the Fig. \ref{Fig_9}.

The ROC plot is drawn FAR vs. GAR, with varying threshold values for YALE database also and is shown in Fig. \ref{Fig_9}. From this plot, the threshold that yields the highest GAR corresponding to the lowest FAR need to be selected. It is evident from the graph that the $G_1$ outperform $G_2$ and $G_3$ for this databset.

\begin{figure}
\centering
\includegraphics[width=73mm,scale=0.6]{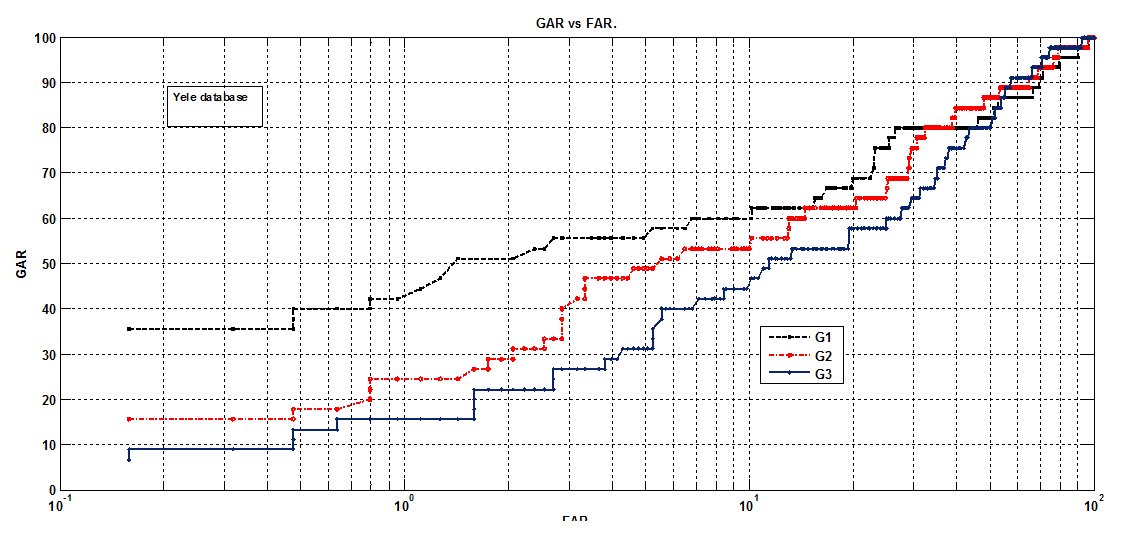}
\caption{The GAR Vs FAR on YALE database with $G_1$, $G_2$ and $G_3$.}
\label{Fig_10}
\end{figure}

\section{COMPARATIVE ANALYSIS}
In order to exhibit the suitability of the proposed approach to face recognition problems, we have made a comparative study with some of the well known and related face recognition approaches. We have considered Local Polynomial Approximation method, Eigenface approach, Local Gabor Binary Pattern Histogram Sequence( LGBP) method, Local Binary Pattern(LBP) method,Fisher and Local Polynomial Approximation-Local Binary Pattern method for comparative analysis. The recognition results on different datasets considering eight number of training images and the remaining number of testing images are shown in the following Table \ref{Tab_10}.

\begin{table}[h]
\caption{Comparative results with existing approaches}
\label{Tab_10}
\begin{tabular}{|l|l|l|ll}
\cline{1-3}
\multicolumn{1}{|l|}{Method} & ORL database & YALE database &  &  \\ \cline{1-3}
Eigenface        & 96.25        & 74            &  &  \\
LGBP             & 96.75        & 86            &  &  \\
LBP				 & 97.5			& 87			&  &  \\
LPA-LBP			 & 99.25		& 92			&  &  \\
MFA				 &92.5			&84				&  &  \\
LPP				 &90			&81				&  &  \\
NPDA			 &90			&82 			&  &  \\
\textbf{Proposed} &             &               &  &  \\
\textbf{method}  &\textbf{97.55}         &\textbf{86.5}           &  &  \\
\cline{1-3}     
\end{tabular}
\end{table}

\section{CONCLUSIONS}
In this work, we have developed Non Binary Local Gradient Contours(NBLGC) as a descriptor for face recognition. A new approach is presented to account for the information from both the central pixel and the neighbourhood pixels of an image while matching test sample with the training samples in the face recognition process thus eliminating the drawback of traditional LBP and LDP approaches which accounts for the neighbouring pixels alone in the representation of texture in a window. We have shown that the newly developed NBLGC descriptor is suitable for face recognition under varying conditions. Since we are losing the information about central pixels and hence we compute the membership based on central pixels. And we are using Shannon entropy for computing the features of the window. Also we have developed a new distance measure which gives most of times better results when compared to conventional Euclidean distance measure. The proposed NBLGC poses good discrimination capability and shown to be worked well with subject having 10 images with 40 classes (ORL) database. As a future work, we are trying to build a classifier and looking for a new membership function.\\
The information value obtained from the window is based on communication theory where the information is defined as the log of the decimal value of a signal. In our work, the information from window is coupled with the concept of information sets in a multiplicative manner. The advantage of this approach is that the information sets allow the information to be modified in different ways. A comparison of performance of the proposed approach is made with that of PCA using two classifiers KNN and SVM. Better results are reported with SVM.\\ 
The proposed approach is found to be effective on images having variation in expression, illumination and pose. The experiments are reported on ORL, Sheffield and Yale face database. The new features are the result of better representation of the uncertainty in the local area like a window, which preserves the information of complete window. The experiments reveal that, better the representation of an uncertainty, better will be the recognition rates.\\

\section*{Authors Biography:}
\textbf{Abdullah Gubbi} is currently working as an Associate Professor in PA College of Engineering, Mangalore. He obtained his Bachelor of Engineering from Gulbarga University, Gulbarga. He received his Masters degree in Electronics  from Walchand College of Engineering  Sangli Maharashtra. His areas of interest include Image Processing, Pattern Recognition and VLSI Design.

\textbf{Mohammad Fazle Azeem}is currently working as an Associate Professor in AMU Aligarh (U.P). He obtained his Bachelor of Engineering from M.M.M. Engineering College, University Of Gorakhpur, Gorakhpur (U.P.) India. He received his Masters degree in Electrical Engineering from Aligarh Muslim University, Aligarh, India. He obtained his Ph.D., from Indian Institute of Technology, Delhi, New Delhi, India. His areas of interest include Control system, Image Processing and VLSI Design.

\textbf{M Sharmila Kumari} obtained her B.E., degree from University of Mysore in the year 2000, M.Tech., degree in Computer Science and Technology from Visweswaraya Technological University in the year 2004 and Ph.D., degree from Mangalore University during 2012. She is currently working as Professor in the Department of Computer Science and Engineering, PA College of Engineering, Mangalore, India. She has authored about forty peer-reviewed papers in International Journals and Conferences. Her areas of research cover Video Processing, Biometrics and Content Based Image Retrieval.

\end{document}